\pdfoutput=1

\documentclass[11pt]{article}

\usepackage{acl}
\usepackage{tabularx}
\usepackage{comment}
\usepackage{placeins}
\usepackage{stfloats} 
\usepackage{placeins}
\usepackage{float} 
\usepackage{tabu} 
\usepackage{multirow}
\usepackage{booktabs}
\usepackage{times}
\usepackage{latexsym}
\usepackage{graphicx}
\usepackage{amsmath}
\usepackage{threeparttable}
\usepackage{xcolor}
\usepackage{enumitem}
\usepackage{subcaption}

\usepackage{tikz}
\usepackage{array}
\usepackage{tabularx}
\usepackage{cellspace}
\usepackage{amsfonts}
\usetikzlibrary{shapes,arrows,positioning}
\definecolor{ao(english)}{rgb}{0.0, 0.5, 0.0}
\definecolor{brass}{rgb}{0.6, 0.8, 0.2}
\definecolor{chromeyellow}{rgb}{1.0, 0.65, 0.0}
\definecolor{crimson}{rgb}{0.86, 0.08, 0.26}

\usepackage{titlesec}

\usepackage{hyperref}

\titlespacing*{\paragraph}{%
  0pt}{
  0.3\baselineskip}{
  1em}

\usepackage{times}
\usepackage{latexsym}
\usepackage{float} 
\usepackage[T1]{fontenc}

\usepackage[utf8]{inputenc}

\usepackage{microtype}

\usepackage{inconsolata}

\newcolumntype{L}[1]{>{\raggedright\let\newline\\\arraybackslash\hspace{0pt}}m{#1}}
\newcolumntype{C}[1]{>{\centering\let\newline\\\arraybackslash\hspace{0pt}}m{#1}}
\newcolumntype{R}[1]{>{\raggedleft\let\newline\\\arraybackslash\hspace{0pt}}m{#1}}

\newcommand{\corpusname}[0]{\textsc{WinoVis}}

\title{Picturing Ambiguity: A Visual Twist on the Winograd Schema Challenge}

\author{%
Brendan Park\textnormal{\thanks{Equal contribution.}, } 
Madeline Janecek\textnormal{\footnotemark[1], }
Naser Ezzati-Jivan\textnormal{,}
Yifeng Li\textnormal{,}
\textnormal{and}
Ali Emami\\
  Brock University, St. Catharines, Ontario, Canada \\
\texttt{\{bp18ul, mj17th, nezzatijivan, yli2, aemami\}@brocku.ca} \\
}

\begin{document}

\maketitle
\begin{abstract}

Large Language Models (LLMs) have demonstrated remarkable success in tasks like the Winograd Schema Challenge (WSC), showcasing advanced textual common-sense reasoning. However, applying this reasoning to multimodal domains, where understanding text \textit{and} images together is essential, remains a substantial challenge. To address this, we introduce \corpusname{},  a novel dataset specifically designed to probe text-to-image models on pronoun disambiguation within multimodal contexts. Utilizing GPT-4 for prompt generation and Diffusion Attentive Attribution Maps (DAAM) for heatmap analysis, we propose a novel evaluation framework that isolates the models' ability in pronoun disambiguation from other visual processing challenges. Evaluation of successive model versions reveals that, despite incremental advancements, Stable Diffusion 2.0 achieves a precision of 56.7\% on \corpusname{}, showing minimal improvement from past iterations and only marginally surpassing random guessing. Further error analysis identifies important areas for future research aimed at advancing text-to-image models in their ability to interpret and interact with the complex visual world.

\end{abstract}

\section{Introduction}
The interpretation of ambiguous constructs in language is crucial for assessing common-sense reasoning, with the Winograd Schema Challenge (WSC) \cite{levesque2011winograd,winograd1972} significantly influencing the evaluation of natural language understanding models. Advances in transformer-based architectures have led Large Language Models (LLMs) to achieve impressive results on WSC-based tasks, approaching near-human performance \cite{brown2020language,sakaguchi2020winogrande,kocijan2023defeat}.

\begin{figure}[]
    \centering
    \includegraphics[width=.95\linewidth]{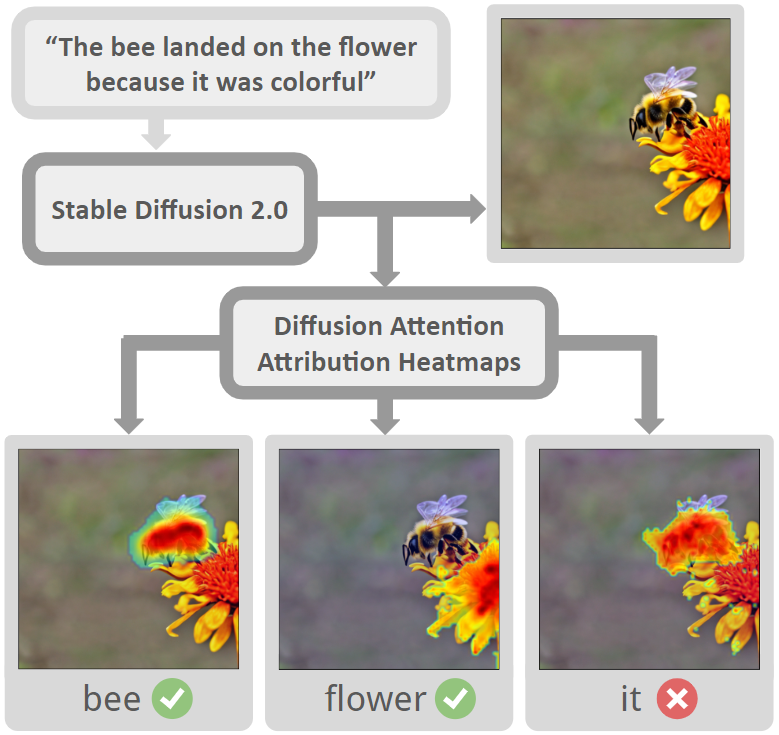}
    \caption{A representative output from Stable Diffusion 2.0 on a \corpusname{} instance. The Diffusion Attentive Attribution Maps (DAAM) clarify the model’s focus for different terms and the correctness of its interpretation: correctly identifying `bee' and `flower' but erroneously associating `it' with the bee instead of the flower.}
    \label{fig:task_overview}
    \vspace{-0.2in}
\end{figure}

Extending common-sense reasoning into multimodal domains, especially disambiguation tasks, is a persisting challenge. Despite the ability of models like Google's Imagen \cite{saharia2022photorealistic}, OpenAI's DALL-E 2 \cite{ramesh2022hierarchical}, and Stability AI's recently open-sourced Stable Diffusion \cite{rombach2022high} to create visually compelling images from text, their interpretability—essential for deciphering the models' reasoning processes—is notably limited \cite{tang-etal-2023-daam}. This gap restricts the development of tools for visuals that match complex texts, reducing model effectiveness when deployed in areas like education and digital media, where text-image integration is essential \cite{dehouche2023s,hattori2023study}.

Our response to this challenge is \corpusname{}, a dataset aimed at probing text-to-image models' common-sense reasoning capabilities through pronoun disambiguation within multimodal scenarios. \corpusname{} not only tests models' ability to distinguish entities within the generated images, but also examines how these models associate pronouns with the correct referents, a nuanced aspect of common-sense reasoning that has been overlooked. As depicted in the \corpusname{} example in Figure \ref{fig:task_overview}, while newer Stable Diffusion models can accurately separate entities within an image, they fail to correctly associate the pronoun `it' with the intended referent, revealing the subtleties and potential gaps in multimodal common-sense reasoning.

The development of \corpusname{} leveraged the generative power of GPT-4 \cite{OpenAI_GPT4_2023,gilardi2023chatgpt}, using a methodical approach to create and refine prompts that elicit common-sense reasoning visually. This process included a complete manual review to ensure each scenario's clarity and relevance for the disambiguation task. Moreover, we introduce a novel evaluation framework that distinguishes between models' pronoun disambiguation proficiency from their handling of visual processing challenges, such as susceptibility to typographic attacks \cite{goh2021multimodal} and semantic entanglement \cite{wu2023uncovering}.

Our contributions are summarized as follows:

\begin{itemize}
\item \textbf{WSC-Adapted Multimodal Dataset (\corpusname{})}: A dataset of 500 scenarios for benchmarking text-to-image models' pronoun disambiguation abilities within a visual context.\footnote{The dataset has been made available at \href{https://github.com/bpark2/WinoVis}{https://github.com/bpark2/WinoVis}.}
\item \textbf{Novel Evaluation Framework for Multimodal Disambiguation}: Metrics and methods designed to isolate pronoun resolution from other visual processing challenges, advancing the understanding of models' common-sense reasoning.
\item \textbf{Insight into Stable Diffusion's Common-Sense Reasoning}: A critical analysis revealing that even state-of-the-art models like Stable Diffusion 2.0 fall significantly short of human-level performance.
\end{itemize}

\section{Background} 
\label{sec:background}
\begin{figure}[!htb]
    \centering
    \includegraphics[width=0.85\linewidth]{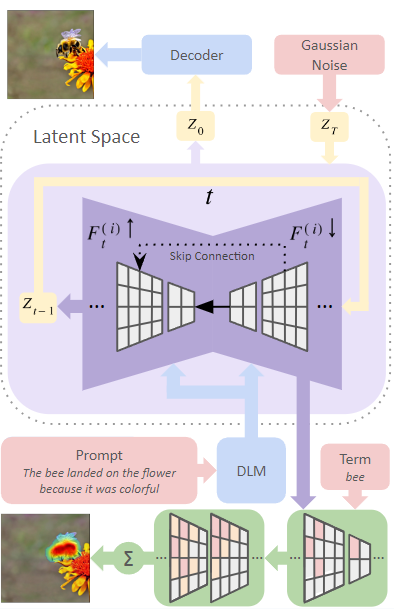}
    \caption{A visual overview of the Stable Diffusion architecture, as well as the Diffusion Attention Attribution Map (DAAM) generation process.}
    \label{fig:daam}
    \vspace{-0.2in}
\end{figure}

\subsection{Latent Diffusion in Image Generation}
\looseness=-1 Latent diffusion models (LDMs) represent a class of generative models designed to synthesize images by progressively refining random noise. A prominent example is Stable Diffusion \cite{rombach2022high}, a text-to-image LDM optimized to generate images from textual prompts. Stable Diffusion integrates three primary components: a deep language model that extracts semantic embeddings from textual prompts; an encoder-decoder architecture for encoding images into latent space representations and decoding them back; and a neural network that is responsible for mean-prediction \cite{Ho2020} (denoted as \(\mu_{\theta}(\boldsymbol{z},\boldsymbol{y},t)\)), noise-prediction \cite{Ho2020} (denoted as \(\epsilon_{\theta}(\boldsymbol{z},\boldsymbol{y},t)\)), or score-prediction \cite{Song2019} (denoted as \(s_{\theta}(\boldsymbol{z},\boldsymbol{y},t)\)). This network is trained on image and text pairs \(\boldsymbol{x}\) and \(\boldsymbol{y}\). During training, which aims to maximize the evidence lower bound (ELBO) \cite{Sohl-Dickstein2015}, the image is initially encoded to \(\boldsymbol{z}_0\), marking the start of the forward diffusion process, formalized as:

{
\vspace{-5mm}
\begin{align*}
    p(\boldsymbol{z}_{t}|\boldsymbol{z}_{t-1}) &= \mathcal{N}(\boldsymbol{z}_{t} | \sqrt{\alpha_{t}} \boldsymbol{z}_{t-1}, (1-\alpha_{t})\boldsymbol{I})\\
    &= \mathcal{N}(\boldsymbol{z}_{t} | \sqrt{1-\beta_{t}} \boldsymbol{z}_{t-1}, \beta_{t}\boldsymbol{I}),
\end{align*}
}
where \(\boldsymbol{z}_{t}\) denotes the latent variable at step \(t\), with \(\beta_{t}=1-\alpha_t\) as the noise schedule hyperparameter, and \(\boldsymbol{I}\) the identity matrix. The U-Net architecture \cite{ronneberger2015unet}, used for denoising, iteratively reverses the diffusion through:

\begin{align*}
&p(\boldsymbol{z}_{t-1}|\boldsymbol{z}_t) = \mathcal{N}\big( \boldsymbol{z}_{t-1} | \mu_{\theta}(\boldsymbol{z}_t,\boldsymbol{y},t), \sigma_t^2 \boldsymbol{I} \big)\\
&=\mathcal{N}\Big( \boldsymbol{z}_{t-1} | \frac{\boldsymbol{z}_t - \frac{\beta_t}{\sqrt{1-\bar{\alpha}_t}} \epsilon_{\theta}(\boldsymbol{z}_t,\boldsymbol{y},t)}{\sqrt{\alpha_t}}, \sigma_t^2 \boldsymbol{I} \Big)
\end{align*}
with \(\sigma_t^2\) as the reverse process noise variance. Cross-attention in the U-Net layers aligns \(\boldsymbol{z}_t\) with \(\boldsymbol{y}\). For conditional generation, the process starts with Gaussian noise \(\boldsymbol{z}_{T}\), conditioned on text \(\boldsymbol{y}\), and refines through reverse diffusion, resembling Langevin dynamics \cite{welling2011bayesian}. For instance, using the score function, we have
\begin{align*}
\boldsymbol{z}_{t-1}^{(j)}=\boldsymbol{z}_{t-1}^{(j-1)} + \frac{\alpha_t}{2} s_{\theta}(\boldsymbol{z}_{t-1}^{(j-1)},\boldsymbol{y},t) + \sqrt{\alpha_t} \varepsilon_j,
\end{align*}
where \(j=1,\ldots,J\), \(J\) is the number of Langevin steps, \(\boldsymbol{z}_{t-1}^{(0)}=\boldsymbol{z}_{t}\), \(\boldsymbol{z}_{t-1}=\boldsymbol{z}_{t-1}^{(J)}\), and \(\varepsilon_j \sim \mathcal{N}(\boldsymbol{0},\boldsymbol{I})\). The denoised \(\boldsymbol{z}_0\) generates the final image, such as the one exemplified in Figure \ref{fig:daam} given a \corpusname{} instance.

\subsection{Diffusion Attentive Attribution Maps}
The Diffusion Attentive Attribution Map (DAAM) technique facilitates interpretability of the influence that different tokens in a prompt have on the image generated by Stable Diffusion models \cite{tang-etal-2023-daam}. This approach capitalizes on the multi-head cross-attention mechanism \cite{vaswani2023attention}, aggregating attention scores from both downsampling and upsampling stages within the U-Net architecture. The attention scores, denoted as \(F_{t}^{(i)\downarrow}\) for downsampling and \(F_{t}^{(i)\uparrow}\) for upsampling, link specific words from the prompt to image regions, signified by coordinates $(x,y)$, across different heads (\(i\)) and layers (\(l\)).

To synthesize a comprehensive heatmap from these attention scores, DAAM applies a spatial normalization procedure, scaling the attention scores for the $k$-th word to match the original image size and summing them across all attention heads ($i$), layers ($l$), and time steps ($t$):

\begin{equation*}
    D_{k}[x,y] = \sum_{i,t,l} \left( F_{t}^{(i)\downarrow}[x,y,l,k] + F_{t}^{(i)\uparrow}[x,y,l,k] \right)
\end{equation*}where  \(F_{t}^{(i)\downarrow}[x,y,l,k]\) and \( F_{t}^{(i)\uparrow}[x,y,l,k]\) represent the bicubically upscaled attention scores for the downsampling and upsampling pathways, respectively. 

DAAM can therefore offer a visual method to evaluate how Stable Diffusion performs pronoun disambiguation, by illustrating where the model concentrates its attention in relation to textual prompts. By examining these visualizations, as demonstrated in Figure \ref{fig:task_overview}, we can discern the model's implicit strategies for linking pronouns with their correct referents.

\begin{table*}
    \footnotesize
    \begin{center}
    \begin{tabu}to\linewidth{@{}X[l,0.22]X[l]X[l,3.7]@{}} 
    \toprule
    
    WSV & Disparate Entities & The \textbf{thief} stole the \textbf{diamond} because \textbf{it} was valuable. ($A =$ diamond)
    \\
    
    & Distinct Entities (Age) & The \textbf{man} carried the \textbf{child} because \textbf{he} was tired. ($A =$ child)
     \\
    
    & Distinct Entities (Role) & The \textbf{king} banished the \textbf{jester} because \textbf{he} was annoying. ($A =$ jester)
     \\
    
    \midrule
    
    WSC & Visually Ambiguous &\textbf{Pete} envies \textbf{Martin} because \textbf{he} is very successful. ($A = $ Martin) \\
    & Entity Exclusion & \textbf{Jane} knocked on \textbf{Susan}'s door, but there was no answer. \textbf{She} was out. ($A = $ Susan) \\ 
    
    \midrule
    
    Filtered & Textually Ambiguous & The \textbf{dog} could not catch the \textbf{squirrel} because \textbf{it} was small. ($A = $ ?) \\
    
    & Illogical & The \textbf{fisherman} cast the \textbf{net} because \textbf{it} was full of holes. ($A = $ net) \\
    
    & Visually Indistinctive & The \textbf{wrestler} defeated the \textbf{opponent} because \textbf{he} was weak. ($A =$ enemy) \\

    & Redundant Entries & \textbf{Anthony} admired \textbf{James} because \textbf{he} was talented. ($A = $ James) \\
    & & \textbf{Ryan} respected \textbf{Andrew} because \textbf{he} was talented.  ($A = $ Andrew)\\

    \bottomrule
    \end{tabu}
    \caption{Examples from the \corpusname{} (WSV) dataset alongside instances from the Winograd Schema Challenge (WSC) dataset and those excluded through manual filtering. In each case, the correct entity is denoted by $A$.}
    \label{tab:items}
    \end{center}
    \vskip -.1in
\end{table*}

\section{Constructing \corpusname{}}

In this section, we detail the methodology behind the creation of \corpusname{},  a novel dataset engineered to assess the pronoun disambiguation capabilities of text-to-image models. The integration of GPT-4 \cite{OpenAI_GPT4_2023,gilardi2023chatgpt} into our dataset generation workflow allowed for significant streamlining of the creation process, achieving reductions in both cost and time, enhancing reproducibility, and reducing the incidence of human error. Our Corpus Construction Cycle consists of two main stages: 1) The GPT Prompt Cycle; and 2) The Manual Filter Process. A full visualization of the process is provided in Appendix Figure \ref{fig:corpus_construction}.

\subsection{Corpus Construction Cycle}
\label{subsec:corpus_construction}
\paragraph{Step 1: GPT Prompt Cycle}

In developing \corpusname{}, we aimed to adapt the Winograd Schema Challenge (WSC) for visual interpretation. This required avoiding the creation of instances that were visually ambiguous, lacked clear visual contexts, or logically didn't necessitate both entities. Table \ref{tab:items} showcases problematic examples from the WSC alongside those of \corpusname{}.

Our iterative prompting process with GPT-4, as outlined in Appendix Table \ref{tab:prompt}, included both successful and problematic few-shot examples to refine the desired outcomes. This approach, detailed entirely in Appendix Table \ref{tab:invalid_instances}, helped clarify what constitutes an acceptable instance. To enhance logical reasoning, we employed a Chain-of-Thought (CoT) \cite{wei2023chainofthought} strategy, further described in Appendix Table \ref{tab:prompt} under \textit{CoT}. Querying instances in batches of ten ensured a varied yet coherent collection while minimizing duplicates.

\paragraph{Step 2: Manual Filter Process} 
After GPT-4 generated the initial set of instances, a manual review was conducted to filter out instances that failed to meet our study's criteria:

\begin{itemize}
    \item \textbf{Textual Ambiguity:} If a prompt could not be easily disambiguated by all annotators it was excluded. 
    \item \textbf{Illogical Content:} Removed if containing nonsensical or irrelevant information.
    \item \textbf{Visual Indistinctiveness:} Omitted when entities lacked clear visual differentiation, essential for accurate entity-pronoun association.
    \item \textbf{Redundancy:} To ensure a broad range of scenarios, instances that were too similar in content or structure were excluded.
\end{itemize}

This manual filtering ensured that each prompt included in \corpusname{} is well suited for evaluating text-to-image models. Examples of excluded instances for each criterion are provided in Table \ref{tab:items}. This review cycle was repeated, refining the selection until achieving a diverse and quality set.

\subsection{Dataset Characteristics}
\label{sec:datchar}
Each sample of \corpusname{} contains a pronoun resolution prompt, a specification of the ambiguous pronoun, an excerpt containing the pronoun, the two referent entity options, the correct referent, and a justification for why the correct entity should be associated with the pronoun.

\paragraph{Disparate and Distinct Entities} The instances within \corpusname{} fall into two broad categories: \textit{Disparate Entities} and \textit{Distinct Entities}. Disparate Entities encompass scenarios with significantly different subjects, such as those across species or object classes (e.g., a person vs. a dog, or a car vs. a tree). Distinct Entities, while sharing some similarities, are visually distinguishable by attributes like age (a mother and child), role (a cop and a thief), or other descriptors, posing more nuanced challenges for pronoun resolution.

\corpusname{} is primarily designed to evaluate a model's \textit{common-sense} reasoning capabilities, rather than to pose a significant challenge. Consequently, a substantial portion of its instances (84.2\%) involve disparate entities. To assess the model in a more demanding context, the remaining 15.8\% of the instances feature distinct entities.

\paragraph{Context Types} To further examine the comprehensiveness of \corpusname{} as a commonsense reasoning benchmark, we categorized each prompt based on the contextual details it provides to link the correct referent to the pronoun. The four contextual categories present in \corpusname{} are:

\begin{itemize}
    \item \textbf{Visually Tangible:} These entries contain descriptions that should have a clear visual impact on the associated referent.
    \item \textbf{Emotional:} These entries describe the emotional or mental state of the referent, which, although more subtle, would still affect the referent's appearance. 
    \item \textbf{Characteristic:} These entries include details pertaining to a referent's personality or nature. While less visually tangible, these details may affect the associated referent's finer details.
    \item \textbf{Visually Intangible:}  These entries involve attributes with minimal to no visual impact on the referent, such as taste, speed, or sound. These entries assess the model’s understanding of purely textual input.
\end{itemize}

We argue that proficiency in pronoun disambiguation requires the capacity to effectively leverage all four context types. Therefore, \corpusname{} includes prompts from each category, providing a comprehensive assessment of a model’s capabilities. Examples and the distribution of each category within \corpusname{} are shown in Appendix Table \ref{tab:context}.

\section{Evaluating Pronoun Disambiguation in \corpusname{}}
\label{sec:pron_disamb_eval}

This section outlines our systematic pipeline to evaluate the capability of Stable  Diffusion models to accurately disambiguate pronouns within the context of \corpusname{}. Our pipeline comprises four stages: 1) Filtering out captioned images to remove visual representations that include embedded text; 2) Enhancing the clarity of distributed attention attribution maps through noise reduction; 3) Excluding images with significant heatmap overlap between referent entities from our analysis; and 4) Determining the model’s final pronoun association by establishing a decision boundary.

\paragraph{Step 1: Caption Filtering}
Text-to-image LDMs sometimes generate images where prompt text appears visually, resulting in `captioned' images. These images erroneously direct a term's attribution to this text, complicating the assessment of the model's visual common-sense reasoning. An example of this is shown in Appendix Figure \ref{fig:image_captioning}.

We therefore specifically excluded captioned images from the analysis set of a studied model, prioritizing those yielding visuals strictly relevant to common-sense interpretation. This exclusion is based on the premise that visual common-sense reasoning should be assessed purely on the model's ability to interpret and generate relevant visual content, without the confounding influence of embedded text. Data on the frequency of prompts resulting in captioned images is detailed in Table \ref{tab:dataset-split}.
\begin{figure*}[]
    \centering
    \includegraphics[width = .9\textwidth]{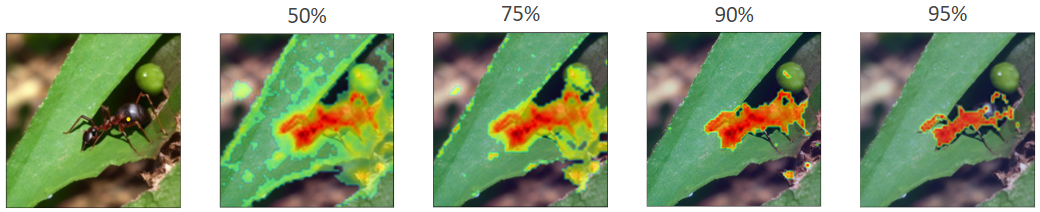}
    \caption{The results of different heatmap thresholds for the prompt ``The ant could not carry the leaf because it was too weak'' and the term `it'.}
    \label{fig:hm_compare}
\end{figure*}

\begin{figure*}
    \centering
    \includegraphics[width = .8\textwidth]{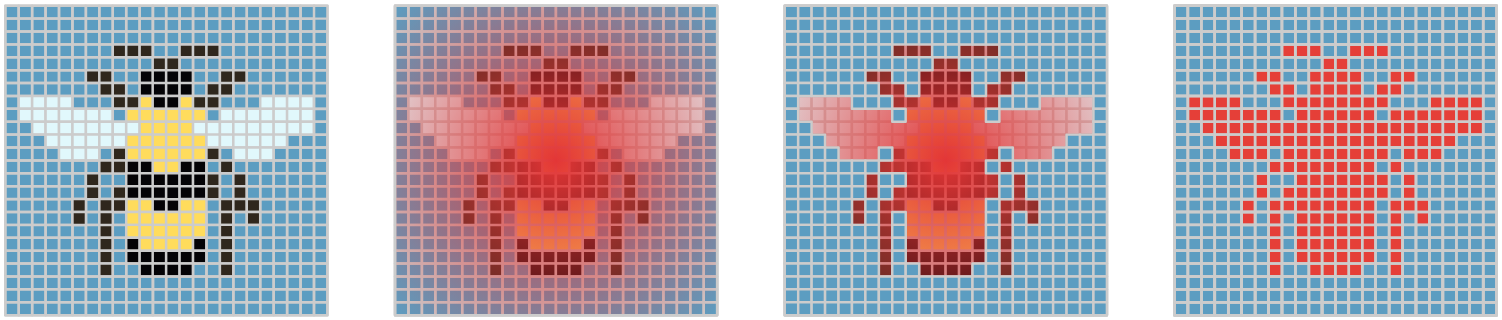}
    \caption{Illustrative example of thresholding on attention maps, progressing through stages to apply a  \(90^{th}\)
  percentile threshold, resulting in a binary mask that accentuates key attention regions.}
    \label{fig:binary-mask}
\end{figure*}

\paragraph{Step 2: Noise Reduction in Attention Maps}

To ensure attention heatmaps clearly reflect the model's focus, we apply a \(90^{th}\) percentile thresholding technique to the heatmaps generated from \corpusname{} prompts. This approach filters out the bottom 90\% of attention scores, considered as noise, and retains only the highest-intensity areas indicative of the model’s primary interest. This \(90^{th}\) percentile threshold was chosen after extensive testing with various thresholds. It was found to be the most effective in balancing the elimination of irrelevant noise while preserving the focal points crucial for understanding the model's interpretation of the prompt. Thresholds below the \(90^{th}\) percentile included too much noise, while higher thresholds risked omitting significant details.

Following this, areas surpassing the threshold are converted into binary masks, delineating significant attention (`1') from the rest (`0'). This representation simplifies the evaluation of the model’s attention distribution, facilitating a more straightforward comparison of its responses to various prompts, thus setting a clearer stage for analyzing how the model associates pronouns with their referents. The impact of this thresholding and the utility of binary masks in enhancing map interpretability are visualized in Figures \ref{fig:hm_compare} and \ref{fig:binary-mask}, respectively.

\paragraph{Step 3: Heatmap Overlap Filtering}

Building on the binary masks created from the previous step, we next employ the Intersection over Union (IoU) metric to further dissect the model's pronoun disambiguation capabilities. The IoU metric, widely recognized in computer vision for evaluating object detection accuracy \cite{he2017mask,szeliski2022computer,takikawa2019gated,zhu2019improving}, measures the overlap between two areas. It is commonly applied to assess the precision of detected objects against ground truth, by comparing their respective binary masks. The IoU calculation is as follows:
\[
\text{IoU} = \frac{\text{Area of Overlap between the binary masks}}{\text{Area of Union of the binary masks}}
\]
This yields a value from 0 (no overlap) to 1 (complete overlap), indicating the strength of association between two terms. 

\begin{figure*}[]
   \centering
   \includegraphics[width=.75\linewidth]{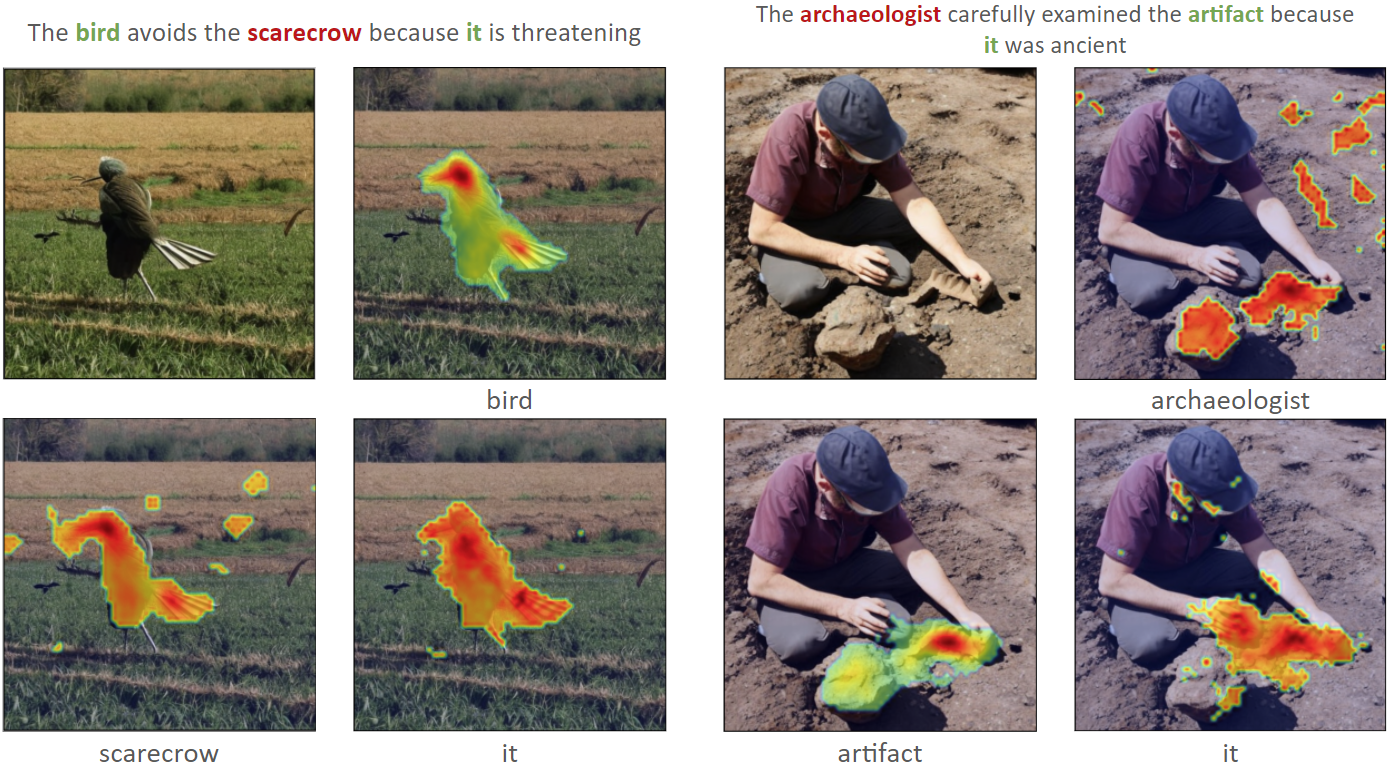}
   \caption{Instances of heatmap overlap generated by Stable Diffusion 2.0 using the \corpusname{} dataset: On the left, two entities lead to nearly identical heatmaps, while on the right, two visually distinct entities show significant heatmap overlap.}
   \label{fig:entangled}
\end{figure*}

For our purposes, a high IoU score between a pronoun and an entity suggests a correct pronoun-to-entity linkage by the model, while high scores between both entities indicate `heatmap overlap'—a state where the model fails to distinguish entity associations, leading to potential misattribution of the pronoun. Refer to Figure \ref{fig:entangled} for examples of this phenomenon.

\begin{figure}
    \centering
    \includegraphics[width=0.9\linewidth]{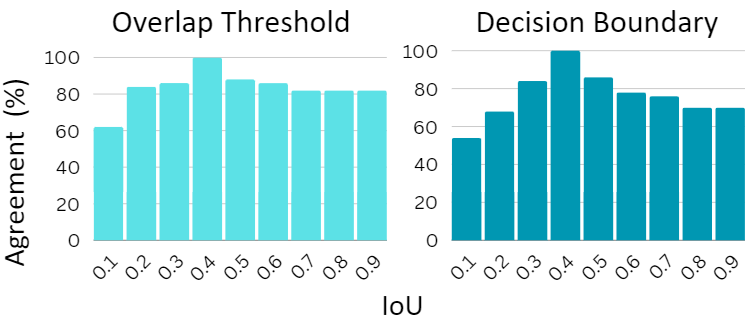}
    \caption{Depicts the level of agreement between the manual decisions and different IoU values for the overlap threshold (left) and decision boundary (right).}
    \label{fig:iouagreement}
\end{figure}

Heatmap overlap complicates pronoun disambiguation, as it reflects a failure to distinguish between entities in the first place. To identify an optimal overlap threshold for detecting such errors, we manually inspected  50 \corpusname{} instances, evaluating heatmap overlays from Stable Diffusion 2.0. A consensus emerged favoring an IoU threshold of 0.4, which yielded full agreement with classifications made by our team, as depicted in Figure~\ref{fig:iouagreement}.

\corpusname{} instances with entity pairs with IoU scores exceeding this threshold are therefore considered invalid, warranting exclusion from further analysis to ensure a focus on clear cases of pronoun disambiguation. This filtering process's impact on the dataset, segmented by model versions is detailed in Table \ref{tab:dataset-split}.

\begin{table}[t]
    \begin{center}
    \begin{tabu}to\linewidth{@{}X[2,c]X[3,c]X[3,c]X[3,c]@{}}
    \toprule
     Version           &  \centering Captioned & \centering Overlapped & \centering \textbf{Evaluable} \\ \midrule
    1.0 & \centering 178 & \centering 24 & \centering 298 \\
    1.5 & \centering 135 & \centering 36 & \centering 329\\
    2.0 & \centering 160 & \centering 71 & \centering 269\\
    XL & \centering 2 & \centering 73 & \centering 425\\
    \bottomrule
    \end{tabu}
    \caption{The number of images generated by Stable Diffusion versions from \corpusname{} prompts, categorized by suitability for pronoun disambiguation analysis.}
    \label{tab:dataset-split}
    \end{center}
\end{table}

\paragraph{Step 4: Making the Final Decision}
In this final step, we utilize the IoU metric once more to establish a decision boundary for evaluating the model's proficiency in pronoun disambiguation. This process involved another comparative analysis conducted by our team of annotators, who manually assessed 50 images generated by SD 2.0 from \corpusname{} instances. Each image was reviewed with its corresponding heatmap to determine the presence of a definitive pronoun-to-entity association. Remarkably, the IoU threshold that aligned with manual assessments was identified again at 0.4, mirroring the overlap threshold. This consistency underscores the threshold's robustness in distinguishing between clear and ambiguous entity associations (Figure \ref{fig:iouagreement} illustrates this agreement).

An IoU score exceeding this threshold signals a strong association between the pronoun and a specific entity, as interpreted by the model. This scenario unfolds in two ways:
\begin{itemize}
    \item If \textbf{only one} referent entity's IoU score with the pronoun surpasses this threshold, it directly informs the model's prediction, indicating a clear pronoun-to-entity association.
    \item If \textbf{both} referent entities' IoU scores exceed the threshold, the entity with the higher score is considered the model's chosen referent. 
\end{itemize}
Predictions are categorized as either \textit{correct} or \textit{incorrect} based on their alignment with the \corpusname{} instance's intended meaning. Cases where neither entity meets the IoU criterion are labeled as \textit{neither}, suggesting the model's failure to disambiguate the pronoun altogether.



    \begin{table*}[t]
    \begin{center}
    \begin{tabu}to\linewidth{@{}X[1.35,c]X[0.75,c]X[0.75,c]X[0.75,c]X[,c]X[,c]X[,c]X[,c]@{}}
    \toprule
    \centering SD Version  & \centering \#Correct & \centering \#Incorrect & \centering \#Neither    & \centering Precision & \centering Recall  & \centering F1-Score &  \centering Certainty \\ \midrule

    1.0  & \centering 24 & \centering 24 & \centering 250 & \centering 50.0 & \centering 8.8 & \centering 14.9 & \centering 16.1 \\    

    1.5  & \centering 38 & \centering 31 & \centering 260 & \centering 55.1  & \centering 12.8 & \centering 20.7 & \centering  21.0 \\

    2.0  & \centering 55 & \centering 42 & \centering 172 & \centering \textbf{56.7} & \centering \textbf{24.2}* & \centering  \textbf{34.1}* & \centering \textbf{36.1}* \\

    XL & \centering 1 & \centering 0 & \centering 424 & \centering N/A & \centering N/A & \centering N/A & \centering 0.24 \\
    
    \bottomrule
    \end{tabu}
    \caption{Comparative performance of Stable Diffusion (SD) models 1.0, 1.5, 2.0, and Stable Diffusion XL (SDXL) \cite{podell2023sdxl} on the \corpusname{} dataset. Metrics are presented as percentages, with * indicating a statistically significant difference for best model (2.0) from second best (1.5) based on a Z-test for two independent proportions (p < 0.01).}
    \label{tab:modelperf}
    \end{center}
\end{table*}

\begin{figure*}
    \centering
    \includegraphics[width=0.85\linewidth]{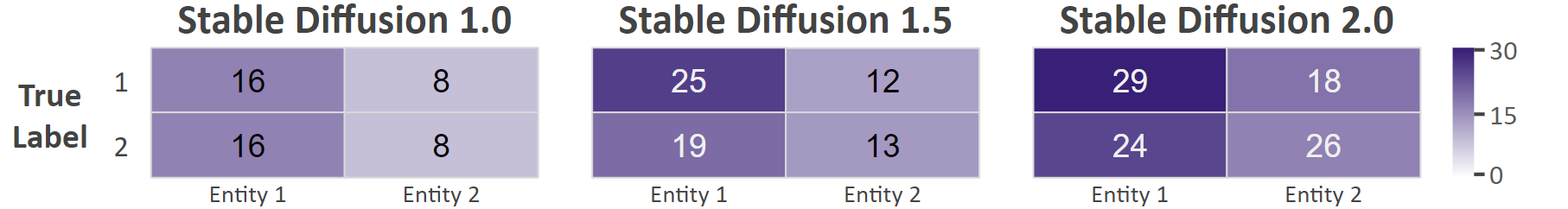}
    \caption{Confusion matrices showing raw count performances of Stable Diffusion models on \corpusname{}. Each matrix provides the counts of predictions for Entity 1 and Entity 2 against their true labels.}
    \label{fig:performance2}
\end{figure*}

\section{Experiments \& Results}

\subsection{Experimental Setup}

\paragraph{Dataset Generation and GPT-4 Configuration} 
For dataset generation, we used GPT-4 (gpt-4-0613; \cite{openai2023gpt4}) with temperature and nucleus sampling (\emph{top-p}) settings optimized to enhance output diversity while adhering to the specific task structure detailed in the prompts. After evaluating temperature values within the range [0,2] with a fixed top-p of 1.0, we determined a temperature of 0.8 as the optimal balance for maintaining both dataset integrity and diversity. The \corpusname{} images were then generated using Stable Diffusion (SD) versions 1.1, 1.5, 2.0 and Stable Diffusion XL (SDXL), through HuggingFace's Diffusers library \cite{von-platen-etal-2022-diffusers}, with each model configured to use 50 diffusion steps.

\paragraph{Diffusion Steps Analysis}
An analysis of image generation quality across different diffusion step settings (20, 50, and 100 steps) was performed to identify the optimal configuration for producing \corpusname{} images. The evaluation criteria included image quality and the presence of unintended captioning. Fifty steps were found to provide the best balance between image quality and computational efficiency, with no significant quality improvements observed at 100 steps.

\paragraph{Main Experiments} 
Using the \corpusname{} dataset, we prompted SD versions 1.1, 1.5, 2.0 and XL to generate corresponding images. Throughout this process, heatmaps for both entities and the pronoun were extracted.\footnote{The heatmap extraction method is based on code available under the MIT License at \href{https://github.com/castorini/daam}{https://github.com/castorini/daam}.} These prepared heatmaps enabled the application of the IoU metric, as elaborated in Steps 3 and 4 of Section \ref{sec:pron_disamb_eval}.

\paragraph{Evaluation Metrics}

We measure model performance using the following metrics, adapted for pronoun disambiguation tasks:

\begin{itemize}
    \item \textbf{Certainty}: The frequency with which the model makes a clear pronoun-to-entity association as opposed to its assocations being marked as `neither'.
    
    \item \textbf{Precision}: The proportion of the model's pronoun-to-entity associations that are correct out of all associations made.
    
    \item \textbf{Recall}: The model's ability to correctly associate pronouns with entities, where `neither' responses are treated as missed opportunities for correct associations (i.e., false negatives).\footnote{An alternative evaluation approach, not penalizing models for ``neither'' predictions, treats the problem as a multi-class classification. Metrics are computed for each entity class and then averaged. For this analysis, see Appendix Table \ref{tab:multiclassperf}.}
    
    \item \textbf{F1-Score}: The harmonic mean of precision and recall, providing an overall measure of the model's disambiguation performance.
\end{itemize}


\subsection{Results}

Table \ref{tab:modelperf} presents the performance of models on the \corpusname{} dataset. Key insights include:

\paragraph{Model Progression and Certainty:} SD 2.0 demonstrates superior precision, recall, and F1-scores, alongside a reduced rate of neither predictions, indicating both progress in pronoun disambiguation and decisiveness. Despite advancements, all models still show a \textit{significant} need for development, with persistent challenges highlighted by the notable proportion of `neither' outcomes and modest precision scores. 

The confusion matrices depicted in Figure \ref{fig:performance2} show the raw count performance of models on the \corpusname{} dataset's pronoun disambiguation problems. Notably, the matrices indicate a gradual decrease in the confusion between entities as the model version increases, with SD 2.0 showing a more distinct separation between the two entities. This suggests an improvement in the models' ability to discern between entities over iterations.

\paragraph{Dismal SDXL Performance:} SDXL's attention maps almost always did not meet the IoU threshold set out for a viable prediction on \corpusname{}. Specifically, the heatmaps attributed to the pronoun were often widely dispersed across the image, resulting in a neither prediction. An example of this can be seen in Appendix Figure \ref{fig:SDXL}. 
    
    The culprit for this may be SDXL's consideration of a large context for high-resolution generation.  Effectively, this may dilute the attention weights of ambiguous tokens and the extra refiner component would impact the generation of attention heatmaps altogether. At the same time, it was intriguing that this issue occurs exclusively for the token corresponding to the ambiguous pronoun (i.e., in Appendix Figure \ref{fig:SDXL}, both the ant and the leaf result in heatmaps that SDXL correctly identifies). This may suggest a tradeoff between  image generation quality and pronoun disambiguation – larger, more capable models  may come with a pronounced cost to interpretability, resulting compromised performance on benchmarks such as WinoVis.

\begin{figure}
    \centering
    \includegraphics[width = 0.84\linewidth]{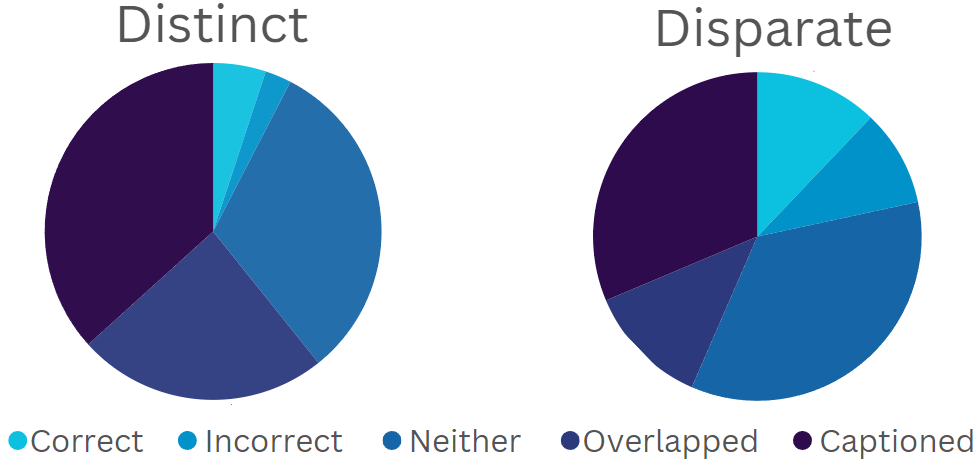}
    \caption{A comparison of the proportion of correct, incorrect, neither, overlapped, and captioned images when SD 2.0 is given distinct versus disparate entities.}
    \label{fig:pie}
\end{figure}

\section{Error Analysis}
In this section, we further examine the performance of the most effective model iteration, SD 2.0. We compare the results across our dataset categories outlined in Section \ref{sec:datchar}, namely disparate and distinct entities. The proportions of correct, incorrect, overlapped, neither, and captioned instances for both categories are visualized in Figure \ref{fig:pie}. 

\begin{figure*}
    \centering
    \includegraphics[width = .75\textwidth]{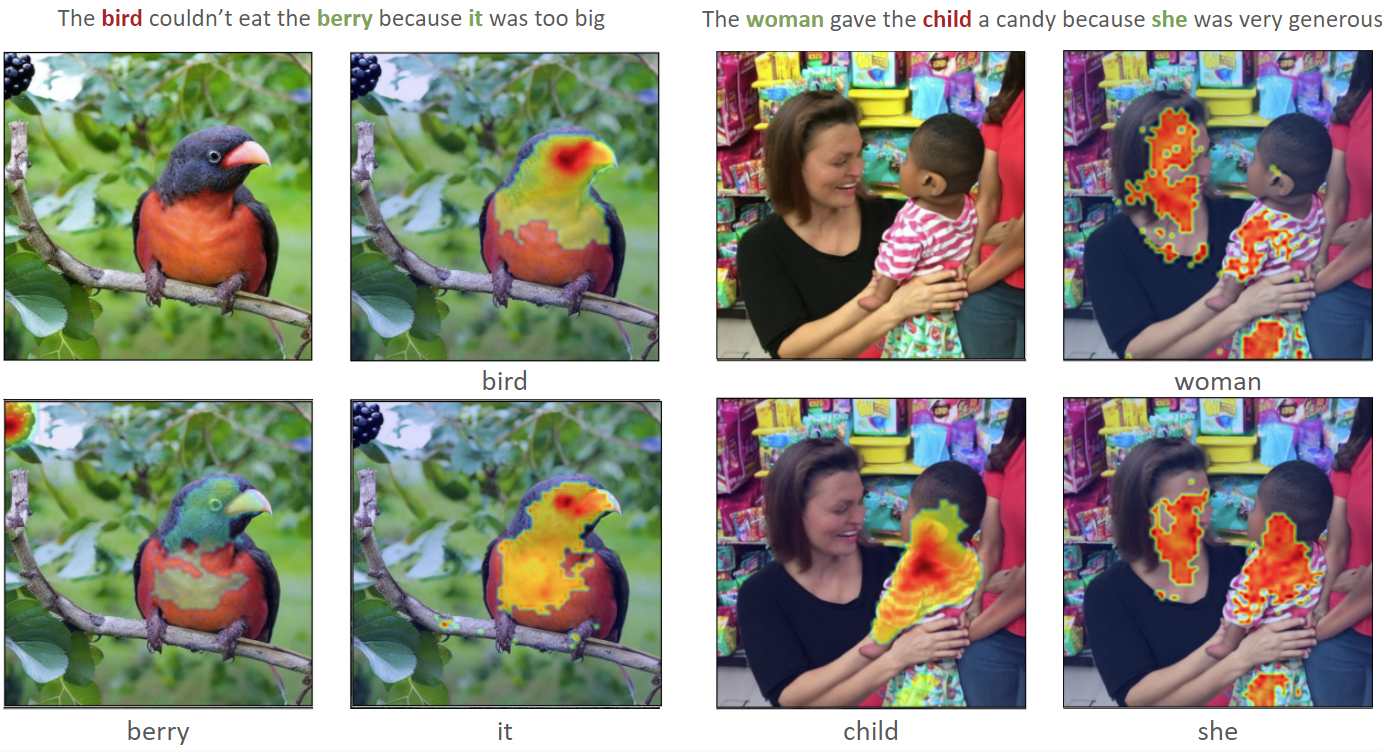}
    \caption{Examples of \textit{incorrect} pronoun associations for \textit{disparate} entities (left) and \textit{distinct} entities (right).}
    \label{fig:pronoun_misinterpretation_examples}
\end{figure*}

\paragraph{Disparate Entities:} In general, SD 2.0 performed the best when working with disparate entities (recall that these were the ``easier'' problems). Over half of the images were evaluable, with the other 43.4\% containing captioning or heatmap overlap. Among the evaluable instances, 31.4\% had neither entity chosen, 9.5\% were incorrect, and 12.1\% were correct. Figure \ref{fig:pronoun_misinterpretation_examples} (left) shows SD 2.0's incorrect pronoun attribution in a \corpusname{} scenario involving disparate entities.

\paragraph{Distinct Entities:} SD 2.0 struggled the most with distinct entities. The majority of instances were not evaluable, with 60.8\% of the items containing captioning or heatmap overlap. Among the evaluable instances, it displayed notable difficulties in making the correct association: in 31.6\% of instances neither entity was chosen, while in 2.5\% of cases, the incorrect entity was chosen. Only 5.1\% of instances resulted in the correct entity being chosen. Figure \ref{fig:pronoun_misinterpretation_examples} (right) depicts an example of two distinct entities, a child and a woman. Interestingly, in this image the pronoun `she' is more strongly attributed to the child instead of the woman, even when the child's gender is not specified.

\section{Related Work}
\paragraph{\textbf{Multimodal Reasoning}} The recent surge in popularity of generative models has underscored the necessity for explainable creativity \cite{llano2020}, leading to a significant body of research investigating the determinants of high-quality prompts for image generation \cite{wang-etal-2023-diffusiondb, oppenleander2023, Pavlichenko2023}. Despite these advancements, the evaluation of how vision models actually interpret prompts is largely underexplored. Most studies focus on the models’ semantic understanding of terms \cite{tang-etal-2023-daam, parcalabescu-etal-2022-valse, thursh2022_winoground} or susceptibility to bias \cite{wang-etal-2023-t2iat}. These evaluations often involve direct, unambiguous prompts, sidestepping more nuanced challenges. \corpusname{} addresses this gap by evaluating the common-sense reasoning of models through the lens of pronoun resolution. This challenge not only expands the scope of assessment for generative models but also sets a new benchmark for understanding their capabilities in interpreting complex linguistic structures. 

\textbf{WSC-Style Tasks}
The Winograd Schema Challenge (WSC) \cite{levesque2011winograd} has catalyzed the development of various datasets aimed at advancing pronominal coreference resolution, each enriching the field by addressing distinct facets of the challenge. Datasets such as Winogrande \cite{sakaguchi2020winogrande} and KnowRef \cite{emami-etal-2019-knowref} expand on the WSC by tackling its limited size, whereas WinoGender \cite{rudinger-etal-2018-gender}, WinoBias \cite{zhao-etal-2018-gender}, and KnowRef-60k \cite{emami2020analysis} study model biases. Further enhancements and crowd-sourcing efforts \cite{wang-etal-2018-glue, trichelair2018on, kocijan-etal-2019-surprisingly, elazar-etal-2021-back, zahraei2024wsc+, sakaguchi2020winogrande} have continually refined the WSC task's scope and methodology.  \corpusname{} uniquely adapts the WSC for text-to-image model evaluation, focusing on multimodal common-sense reasoning. It introduces the challenge of visually disambiguating pronouns, filling a crucial gap in multimodal evaluation.

\section{Conclusion}
\looseness=-1 This paper presented \corpusname{}, a new approach to test how well text-to-image models like Stable Diffusion handle pronoun disambiguation. Our work reveals significant gaps in these models' abilities to interpret ambiguous scenarios accurately. Central to our contribution is a novel evaluation framework designed to isolate common-sense reasoning in pronoun disambiguation from well-studied challenges such as typographic attacks and semantic entanglement. Future research should build on our groundwork to develop models that not only generate visually compelling images but also accurately understand the narratives and relationships within them.

\section*{Limitations}
\paragraph{Entity Separation:} Stable Diffusion models encounter challenges with distinguishing between two semantically similar entities. This can be seen in either heatmap overlap or entanglement, both of which result in a significant proportion of generated images being unsuitable for pronoun disambiguation. Entanglement is particularly pronounced in images generated from prompts featuring semantically similar entities. Since sentences from \corpusname{} often employ such entities to introduce ambiguity, resolving entanglement could improve the model's ability to distinguish individual entities and expand the range of Winograd-like prompts that Stable Diffusion can visualize for our analysis. 

\paragraph{Model Diversity:} Due to its open-source nature, Stable Diffusion facilitated the creation of heatmaps using DAAM, a capability not available in closed-source LDMs. Currently, DAAM is the only framework which enables the interpretation of such models and is specifically designed for Stable Diffusion. Future research should investigate methods to enhance interpretability across a wider range of LDMs and multi-modal diffusion models (and more open-source ones, as they become increasingly available), enabling their assessment in pronoun disambiguation using \corpusname{}.

\paragraph{Bias Analysis:} Our study does not explicitly address potential biases in Stable Diffusion that might influence its decision-making processes. Instances of incorrect pronoun resolution, such as the woman-child example depicted in Figure \ref{fig:pronoun_misinterpretation_examples}, hint at underlying biases. Future work should rigorously explore these biases and their effects on model performance. Investigating whether Stable Diffusion exhibits systematic preferences in resolving ambiguities could uncover patterns in its reasoning strategies, guiding efforts to mitigate biases and enhance multimodal pronoun disambiguation capabilities.

\paragraph{Dataset Diversity:} Although efforts were made to maximize dataset diversity during the generation of samples for \corpusname{}, opportunities for enhancement remain. Further refinement could entail creating samples that exhibit greater complexity and encompass a broader spectrum of circumstances, entities, and instances of ambiguous pronouns.

\paragraph{Filtering Limitations:} Although our filtering process aimed to minimize the impact of model weaknesses on our analysis, exceptions exist. In certain cases, semantic entanglement eluded detection through heatmap overlap measures (see Appendix Figure \ref{fig:filtering_limitation} for an example). Future research should investigate alternative detection methods to better mitigate the influence of such model flaws on our analysis of \corpusname{}.

\section*{Acknowledgements}
This work was supported by the Natural Sciences and Engineering Research Council of Canada and by the New Frontiers in Research Fund.

\bibliography{custom_bp}
\newpage

\clearpage
\onecolumn

\appendix

\section{Appendix}
\label{sec:appendix}

\label{sec:construction_wsv}
\begin{figure}[h]
    \centering
    \includegraphics[width=\textwidth]{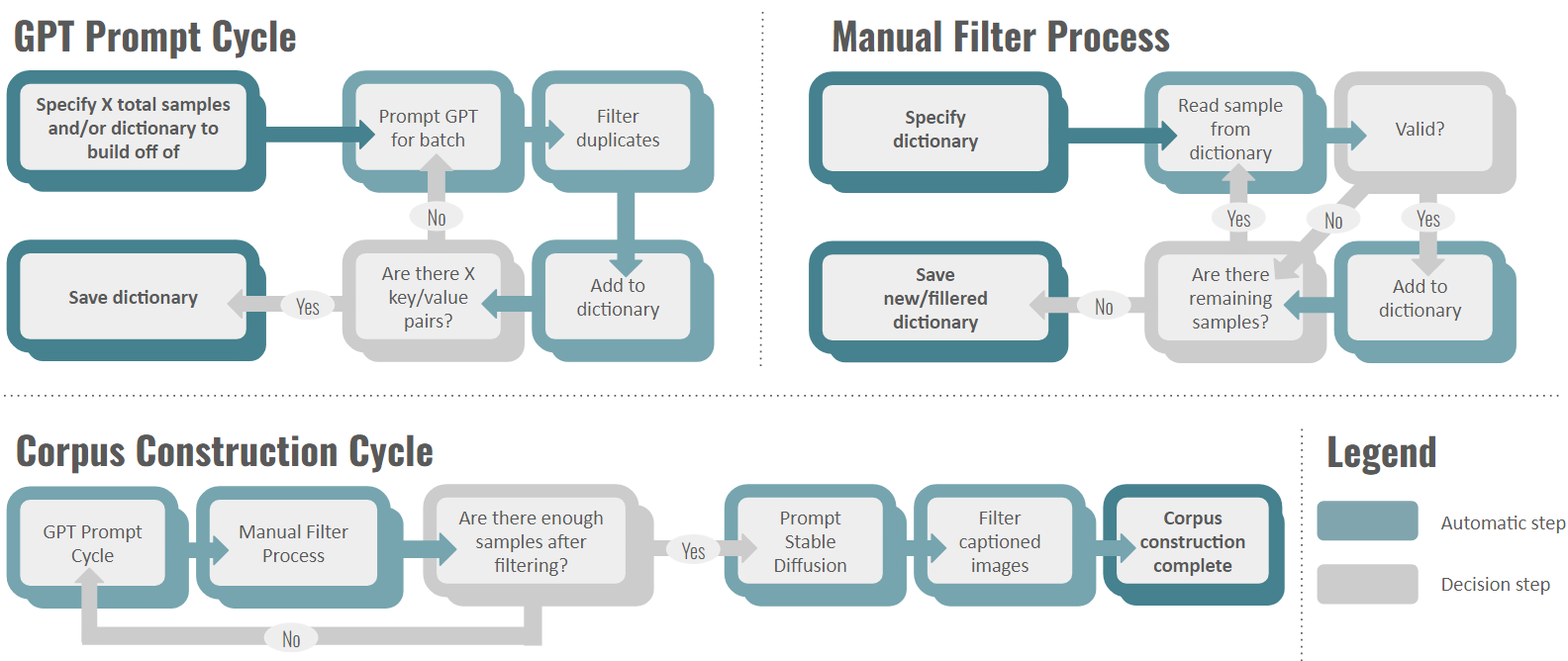}
    \caption{A visual overview of our Corpus Construction Cycle.}
    \label{fig:corpus_construction}
\end{figure}

\begin{figure}[h]
    \centering
    \includegraphics[width=0.95\linewidth]{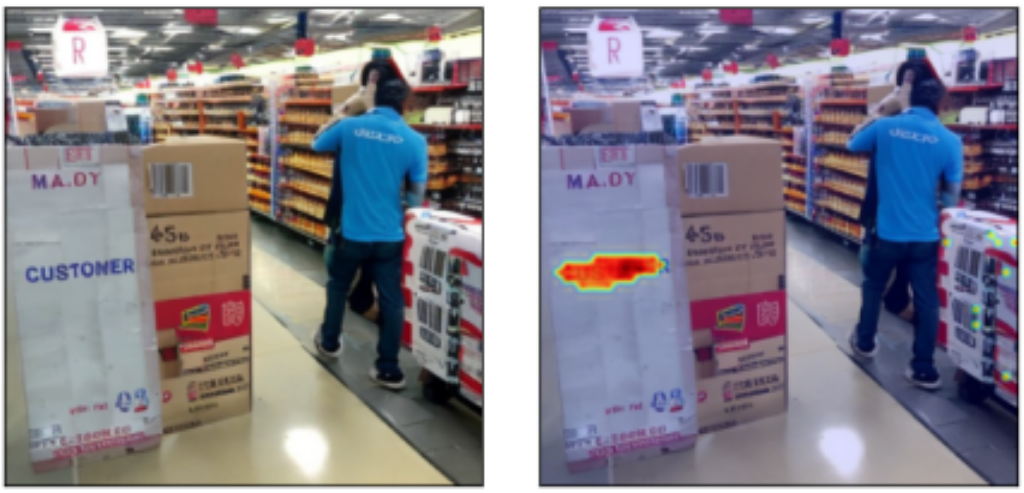}
    \caption{An example of image captioning. In this case, the prompt ``The customer returned the product because it was unsatisfied" produced an image that includes the word `customer'. The attribution heatmap for the term `customer' focuses on this text. }
    \label{fig:image_captioning}
\end{figure}

\begin{table}[h!]
    \begin{center} \centering
    \begin{tabu}to\linewidth{@{}X[1,c]X[1,c]X[1,c]X[1,c]X[1,c]X[,c]X[,c]X[,c]X[,c]@{}}
    \toprule
     \multirow{2}{*}{Model} &   \multicolumn{2}{c}{\underline{Correct Attribution}} &    \multicolumn{2}{c}{\underline{Incorrect Attribution}} & \multirow{2}{*}{Accuracy}  & \multirow{2}{*}{Precision} &  \multirow{2}{*}{Recall}  & \multirow{2}{*}{F1-Score}  \\ 
     
      & Entity 1 & Entity 2 & Entity 1 & Entity 2 \\ 
     \midrule

    1.0  &  16 &  8 & 16 & 8  & 50.0   &  50.0&  50.0& 50.0\\    

    1.5  &  25 & 13 & 19  & 12  & 55.1  &  54.4& 54.1& 54.3 \\

    2.0  &  29 &  26& 24 & 18 & \textbf{56.7} & \textbf{56.9} &\textbf{56.9} & \textbf{56.9}\\
    
    \bottomrule
    \end{tabu}
    \caption{An alternate evaluation of Standard Diffusion models that treats the problem as a multi-class classification task. The reported Precision and Recall scores are computed by taking the average of both entity classes. }
    \label{tab:multiclassperf}
    \end{center}
\end{table}

\begin{table}[]
\begin{tabular}{|c|l|}
\hline
Component & Prompt Content \\ \hline
Setup     & 
        \begin{tabular}[c]{@{}l@{}}
            A Winograd schema sentence is a sentence that contains an ambiguity and requires \\
            world knowledge and reasoning for its resolution. For example: The city councilmen\\
            refused the demonstrators a permit because they feared violence.\\
            
            Here, ``they” presumably refers to the city council; because city councils are\\
            typically responsible for maintaining order and avoiding violence in their city. It\\
            is more plausible that a city council would fear violence than actively advocate for \\
            it. In this example we get the answer based on our world knowledge that tells us city \\
            councils generally wish to preserve order, while protest movements sometimes embrace \\
            confrontation and violence to achieve political aims. This matches the logical \\
            referents in the schema.\\
                    \end{tabular} \\ \hline                                                                        
Criteria     & 
        \begin{tabular}[c]{@{}l@{}}
            Winograd schema sentences must abide by five rules:\\
            1. Be easily disambiguated by the reader;\\
            2. Not be solvable by simple techniques such as selectional restrictions;\\
            3. The ``snippet'' must directly refer to the entity specified by the ``answer''\\
            4. Neither of the ``options'' should be found in the ``snippet''.\\
            5. The ``pronoun'' must be applicable to both ``options''. For example, two men \\
            could share the pronoun ``he'' or ``him''. Furthermore, a person with an occupation\\
            such as an athlete or doctor and a non-human entity cannot share the pronouns ``he'' \\
            or ``she'' but may share ``it''. If a plural pronoun is used such as ``they'' then both \\
            ``options'' should also be plural. For example, coaches instead of coach and players \\
            instead of player. \\
        \end{tabular} \\ \hline
Examples          &                                                                    \begin{tabular}[c]{@{}l@{}}
            Here is an example of some sentences which match the format of the Winograd schema: \\
            (using output with reason examples)  \\
            \textbf{INSERT WSC SAMPLES} \\
            An example of an invalid pair is: \\
            Sentence1:\\\{\\``statement'': ``The boy kicked the ball because it was deflated.'',\\``pronoun'': ``it'',\\``snippet'': ``it was deflated'',\\``options'': [``the boy'', ``the ball''],\\``answer'': 1,\\``reason'': ``If 'deflated' is used, it implies the ball was deflated.''\\\}
        \end{tabular} \\ \hline
CoT     & 
        \begin{tabular}[c]{@{}l@{}}
            Without skipping any, come up with \textbf{BATCH\_SIZE} new valid sentences starting at 
            \\sentence one. Think step by step for each new sentence by following these steps:\\
            1. Come up with two entities or objects which share a pronoun.\\
            2. Think of a pronoun that seems just as semantically compatible with the two \\
            antecedent options, but can be disambiguated using common sense reasoning and not\\
            at all with distributional cues between the antecedents and the rest of the sentence.\\
            3. Come up with a completely new sentence that follows the principles of the example \\
            sentences and follows the rules listed above.\\
            Repeat this process for all the sentences you generate. The sentences should be original \\
            and diverse in the topics that they cover.
        \end{tabular} \\ \hline
\end{tabular}
\caption{The prompt used in the Corpus Construction Cycle broken down into distinct sections. }
\label{tab:prompt}

\end{table}

\begin{table}[] \centering
\begin{tabular}{|l|}
\hline

            The athlete left the game because it was [risky/exhausting].\\
            a:\\
            \{``statement": ``The athlete left the game because it was risky.",\\
            ``pronoun": ``it",\\
            ``snippet": ``it was risky",\\
            ``options": [``athlete", ``game"],\\
            ``answer": 1,\\
            ``reason": ``If `risky' is used, it implies the game was risky, causing the \\
            athlete to leave."\}\\
            b:\\
            \{``statement": ``The athlete left the game because it was exhausting.",\\
            ``pronoun": ``it",\\
            ``snippet": ``it was exhausting",\\
            ``options": [``athlete", ``game"],\\
            ``answer": 0,\\
            ``reason": ``If `exhausting' is used, it implies the athlete was exhausted, \\
            causing him to leave the game."\}\\
            
            Explanation: The ``snippet" refers to the game's impact on the athlete when it\\
            should refer to the ``athelete" itself. To correct this sample, the term used should\\
            be exhausted instead of exhausting.

\\ \hline                                                                        
            The boy kicked the ball because it was [deflated/inflated].\\
            a:\\
            \{
            ``statement": ``The boy kicked the ball because it was deflated.",\\
            ``pronoun": ``it",\\
            ``snippet": ``it was deflated",\\
            ``options": [``the boy", ``the ball"],\\
            ``answer": 1,\\
            ``reason": ``If `deflated' is used, it implies the ball was deflated."
            \}\\
            b:\\
            \{
            ``statement": ``The boy kicked the ball because it was inflated.",\\
            ``pronoun": ``it",\\
            ``snippet": ``it was inflated",\\
            ``options": [``the boy", ``the ball"],\\
            ``answer": 1,\\
            ``reason": ``If `inflated' is used, it implies the ball was inflated, prompting \\the boy to kick it."\}\\
            
            Explanation: In a Pair, a and b must not have the same ``answer". If Pair2.a's \\
            ``answer" is 0, Pair2.b's ``answer" should be 1 and vice-versa.
 \\ \hline
\end{tabular}
\caption{Examples of invalid instances that were included in the prompt used in the Corpus Construction Cycle. }
\label{tab:invalid_instances}

\end{table}

\begin{figure}
    \centering
    \includegraphics[width=\linewidth]{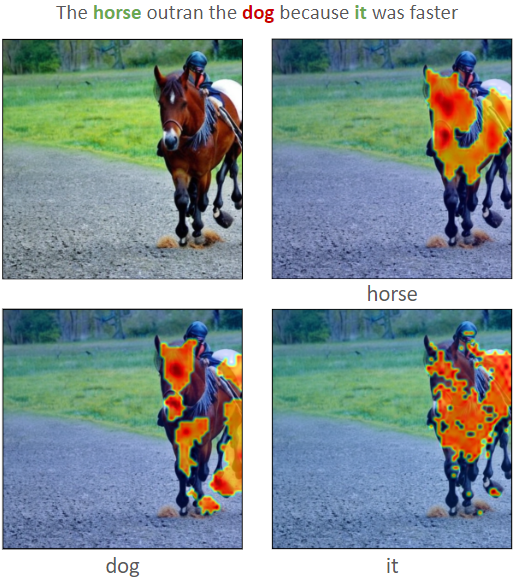}
    \caption{An example of a generated image containing only one of the entities from the prompt. While the horse is visible, the dog is not.}
    \label{fig:entity_exclusion}
\end{figure}

\begin{figure}
    \centering
\includegraphics[width=\linewidth]{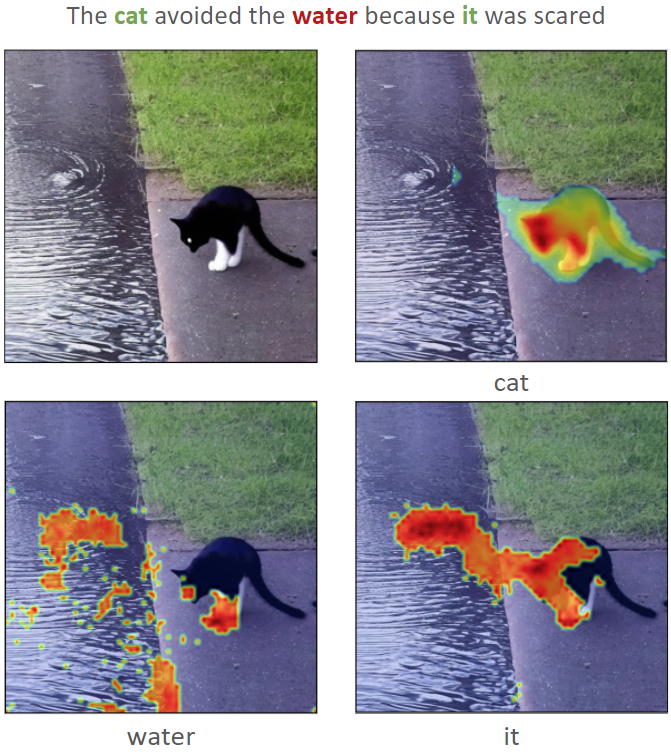}
    \caption{An example of the case where the DAAM heatmap for the pronoun does not clearly indicate a decision made by the model. Rather than overlapping with either the `cat' or the `water', the heatmap for `it' appears to overlap slightly with both while also encompassing some space not seen in either of the entities' heatmaps.}
    \label{fig:uncertainty}
\end{figure}

\begin{figure}
    \centering
\includegraphics[width=\linewidth]{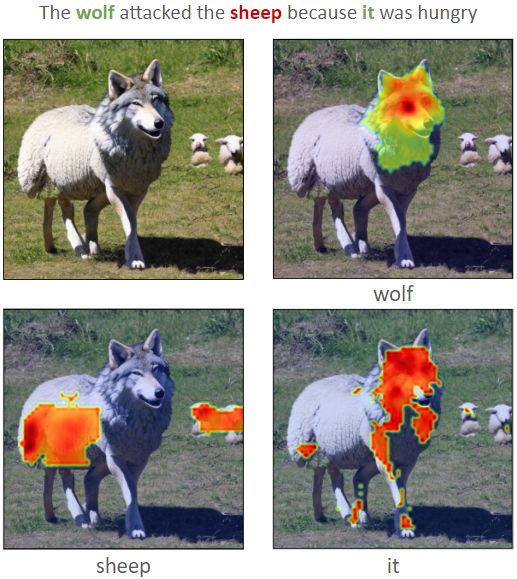}
    \caption{An example of an image that was not automatically filtered via measurement of heatmap overlap. While the two entities are semantically entangled their heatmaps are distinct (non-overlapped).}
    \label{fig:filtering_limitation}
\end{figure}

\begin{figure}
    \centering
\includegraphics[width=\linewidth]{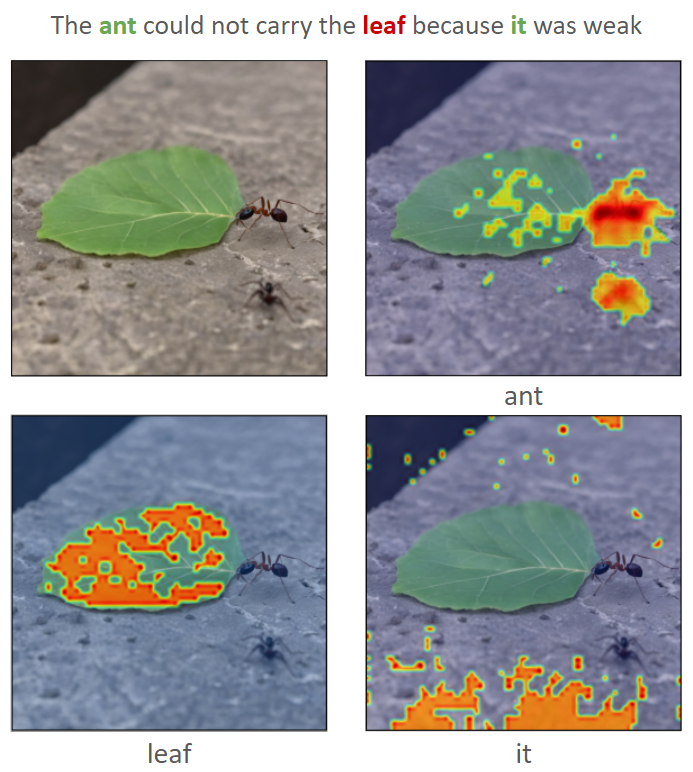}
    \caption{An example of an image generated by SDXL. Here, both entity heatmaps overlap correctly with their respective visual representations. However, the heatmap for the ambiguous pronoun is distributed across the image showing a lack of certainty in the model's decision.}
    \label{fig:SDXL}
\end{figure}

\begin{table*}
    \footnotesize
    \begin{center}
    \begin{tabu}to\linewidth{@{}X[l]X[c]X[l,2.5]@{}} 
    \toprule
    
    Context Type & \% of WSV & Example \\
    \midrule

    Visually Tangible & 38.6 & The \textbf{plumber} had to replace the \textbf{pipe} because \textbf{it} \textit{was rusty}. \\
    
    Emotional & 15.0 & The \textbf{dog} chased the \textbf{car} because \textbf{it} \textit{was excited}. \\

    Characteristic & 29.2 & The \textbf{king} did not trust the \textbf{advisor} because \textbf{he} \textit{was deceitful}. \\

    Visually Intangible & 17.2 & The \textbf{cat} is afraid of the \textbf{vacuum cleaner} because it \textit{is loud}. \\

    \bottomrule
    \end{tabu}
    \caption{Examples taken from the \corpusname{} (WSV) dataset exhibiting the four different context types.}
    \label{tab:context}
    \end{center}
    \vskip -.1in
\end{table*}

\end{document}